\documentclass[conference]{IEEEtran}
\IEEEoverridecommandlockouts
\usepackage{cite}
\usepackage{mathptmx}
\usepackage{amsmath,amssymb,amsfonts}
\usepackage{algorithmic}
\usepackage{graphicx}
\usepackage{textcomp}
\usepackage{setspace}
\usepackage{xcolor}
\usepackage{array}
\def\BibTeX{{\rm B\kern-.05em{\sc i\kern-.025em b}\kern-.08em
    T\kern-.1667em\lower.7ex\hbox{E}\kern-.125emX}}
\begin{document}

\title{Handwritten Text Recognition Using Convolutional Neural Network\\
}

\author{\IEEEauthorblockN{Atman Mishra}
\IEEEauthorblockA{\textit{Dept. of AIML} \\
\textit{New Horizon College of Engineering}\\
Bangalore, India \\
atmanmishra1122@gmail.com}
\and
\IEEEauthorblockN{A. Sharath Ram}
\IEEEauthorblockA{\textit{Dept. of AIML} \\
\textit{New Horizon College of Engineering}\\
Bangalore, India \\
sharath29.ashok@gmail.com}
\and
\IEEEauthorblockN{Kavyashree C.}
\IEEEauthorblockA{\textit{Dept. of AIML} \\
\textit{New Horizon College of Engineering}\\
Bangalore, India \\
kavyashree.csd@gmail.com}
}

\maketitle

\begin{abstract}
OCR (Optical Character Recognition) is a technology that offers comprehensive alphanumeric recognition of handwritten and printed characters at electronic speed by merely scanning the document. Recently, the understanding of visual data has been termed Intelligent Character Recognition (ICR). Intelligent Character Recognition (ICR) is the OCR module that can convert scans of handwritten or printed characters into ASCII text. ASCII data is the standard format for data encoding in electronic communication. ASCII assigns standard numeric values to letters, numeral, symbols, white-spaces and other characters. In more technical terms, OCR is the process of using an electronic device to transform 2-Dimensional textual information into machine-encoded text. Anything that contains text both machine written or handwritten can be scanned either through a scanner or just simply a picture of the text is enough for the recognition system to distinguish the text.

The goal of this papers is to show the results of a Convolutional Neural Network model which has been trained on National Institute of Science and Technology (NIST) dataset containing over a 100,000 images. The network learns from the features extracted from the images and use it to generate the probability of each class to which the picture belongs to. We have achieved an accuracy of 90.54\% with a loss of 2.53\%.
\end{abstract} 


\section{Introduction}
In the past three decades, much work has been devoted to handwritten text recognition, which is used to convert human-readable handwritten language into machine-readable codes. Handwritten text recognition has attracted a great deal of interest because it provides a method for automatically processing enormous quantities of handwritten data in a variety of scientific and business applications. The underlying problem with handwritten text has been that various individuals' representations of the same character are not identical. An additional difficulty experienced while attempting to decipher English handwritten characters is the variance in personal writing styles and situational differences in a person's writing style. In addition, the writer's disposition and writing environment may influence writing styles.

Attempts to create a computer system that could understand handwriting are the only time the complexity of optical pattern recognition becomes apparent. The strategy using artificial neural networks is thought to be the most effective for creating handwriting recognition systems. Neural networks significantly help in modelling how the human brain operates when identifying handwritten language in a more efficient manner. It enables machines to interpret handwriting on par with or better than human ability. Humans use a variety of writing styles, many of which are difficult to read. Additionally, reading handwriting may be time-consuming and difficult, particularly when one is required to examine several documents with handwriting from various persons. Neural networks are the best choice for the suggested system since they can extract meaning from complicated data and spot trends that are hard to spot manually or using other methods. The primary goal of this project is to create a model based on the concept of Convolution Neural Network that can recognize handwritten digits and characters from a picture. We have built a simple Convolutional Neural Network (CNN) system which has been trained on NIST dataset. 

\section{Related Work}
Many researchers tried to develop handwritten text recognition models in the past, but none of them are perfectly accurate and it still requires much more research in this field. 

M. Brisinello et al [1] proposed a pre-processing method which improves Tesseract OCR 4.0’s performance by approximately 20\%. They implemented a two-step process which involves clustering of input images and a classifier which identifies whether the images contain text or not. 

In [2], neural networks are used to sample the pixels in the image into a matrix and match them to a known pixel matrix. It achieved an astounding accuracy of 95.44\%. 

An open-source OCR engine developed by HP called Tesseract OCR Engine has been implemented in this research which has achieved an accuracy of 95.305\% which can be seen in [3]. 

A different approach has been implemented in [4] which has used RNN and embedded system for character recognition of Devanagari which involves 120 Indian languages. 

Methods that were based on confusion matrices were employed in the study that Yuan-XiangLi et al [5] conducted. The approach starts with the original candidates, utilises them to conjecture which characters are most likely to be correct, and then combines the postulated set with the original candidates to obtain a new set of candidate candidates.

In [6] by T. Sari et al., the technology of character segmentation has been incorporated in a different manner of implementation. In many OCR systems, character segmentation is a prerequisite step for character recognition. It is crucial because poorly fragmented characters are unlikely to be correctly recognised.

\section{Convolutional Neural Network}

In the field of Deep Learning, Convolutional neural networks (CNNs) are a type of Artificial Neural Network (ANN) which are frequently used to analyse visual data, including photographs and videos, in order to discover patterns or trends in the visual data.. This type of analysis is typically carried out in order to discover new applications for the visual data. A Multilayer perceptron is typically connected in a fully connected network, and CNN is a simplified version of this type of network architecture. The Convolutional Neural Network (CNN) is made up of a variety of components and operations, such as Pooling, Convolution, and Neural Network, among others. 

\subsection{Neural Networks}

Neural Networks or Artificial Neural Networks are computerized systems which are developed in order to replicate animal brain. ANNs are usually used to perform a certain limited task but they can be trained to perform any kind of difficult tasks which can replicate or are better than task done by humans. 

Neural networks are composed of 3 layers shown in Fig 1
\begin{itemize}
  \item Input Layer
  \item Hidden Layer
  \item Output Layer
\end{itemize}

\begin{figure}[htpb]
\centerline{\includegraphics[scale=.3]{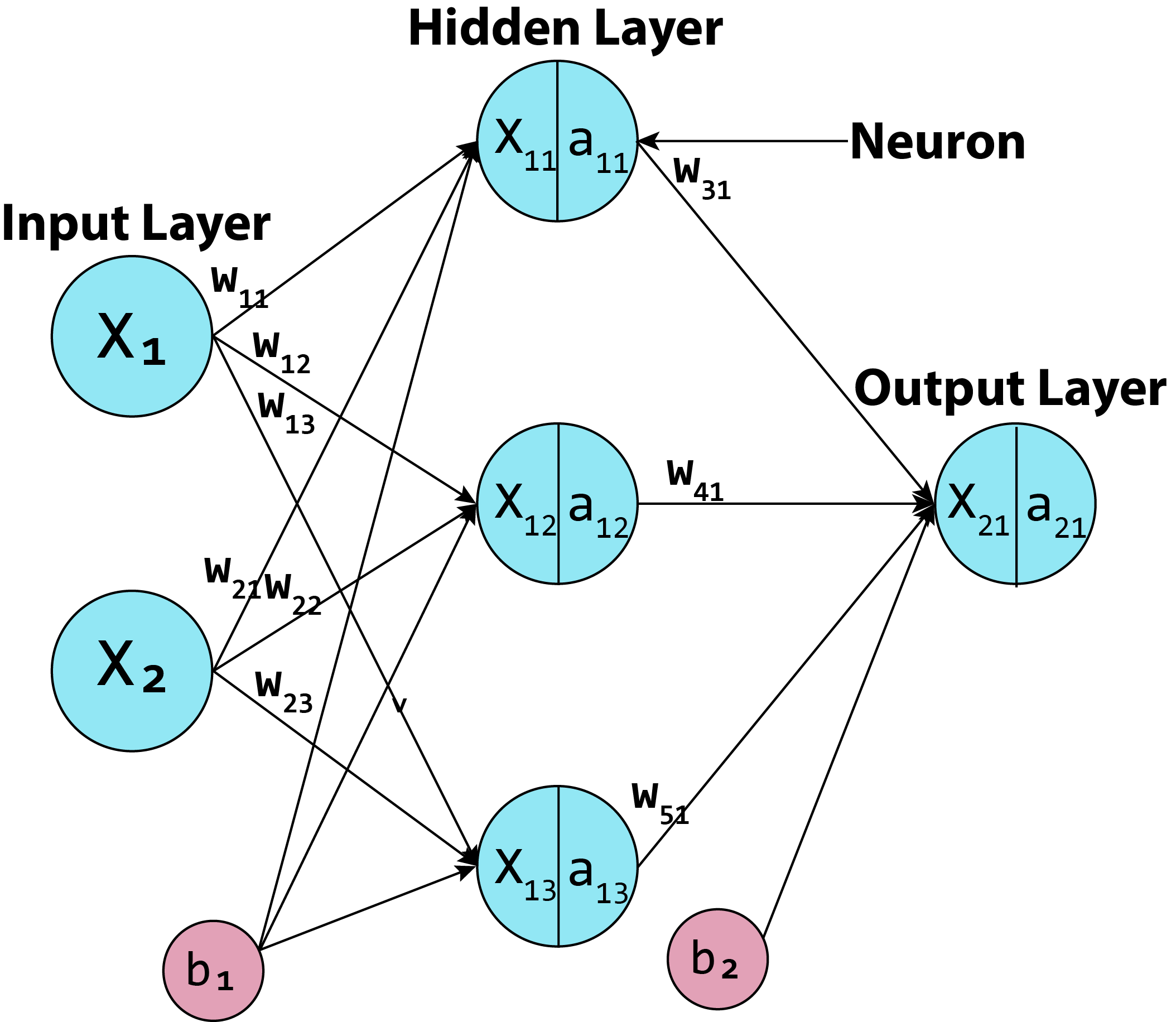}}
\caption{A Basic Neural Network}
\end{figure}Every artificial neuron receives an input and produces a single output that can be spread to a number of different artificial neurons. The feature values of a sample of external data, such as photographs, can serve as the inputs. Alternatively, the outputs of other neurons can serve in this capacity. The objective, such as recognising an object in an image, is completed successfully by the outputs of the final output neurons of the neural network. They are also known as Weighted Graphs rather frequently. Each connection between neurons has a weight 'W' to them along with a bias 'b' added in order to make a weighted input. After that, the weighted input will be utilised in an activation function that is present in a neuron in order to bring about non-linearity in the output of the neuron.

An activation function is a mathematical function employed in the hidden layers that uses the weighted input and bias to determine whether or not the particular neuron will be activated. The activation function also applies a non-linear transformation to the input, which enables the system to acquire new knowledge about the input. The output of the activation function will be regarded as an input to the corresponding neuron.The above mentioned process for a single neuron is given in Fig 2:

\begin{figure}[htpb]
\centerline{\includegraphics[scale = .4]{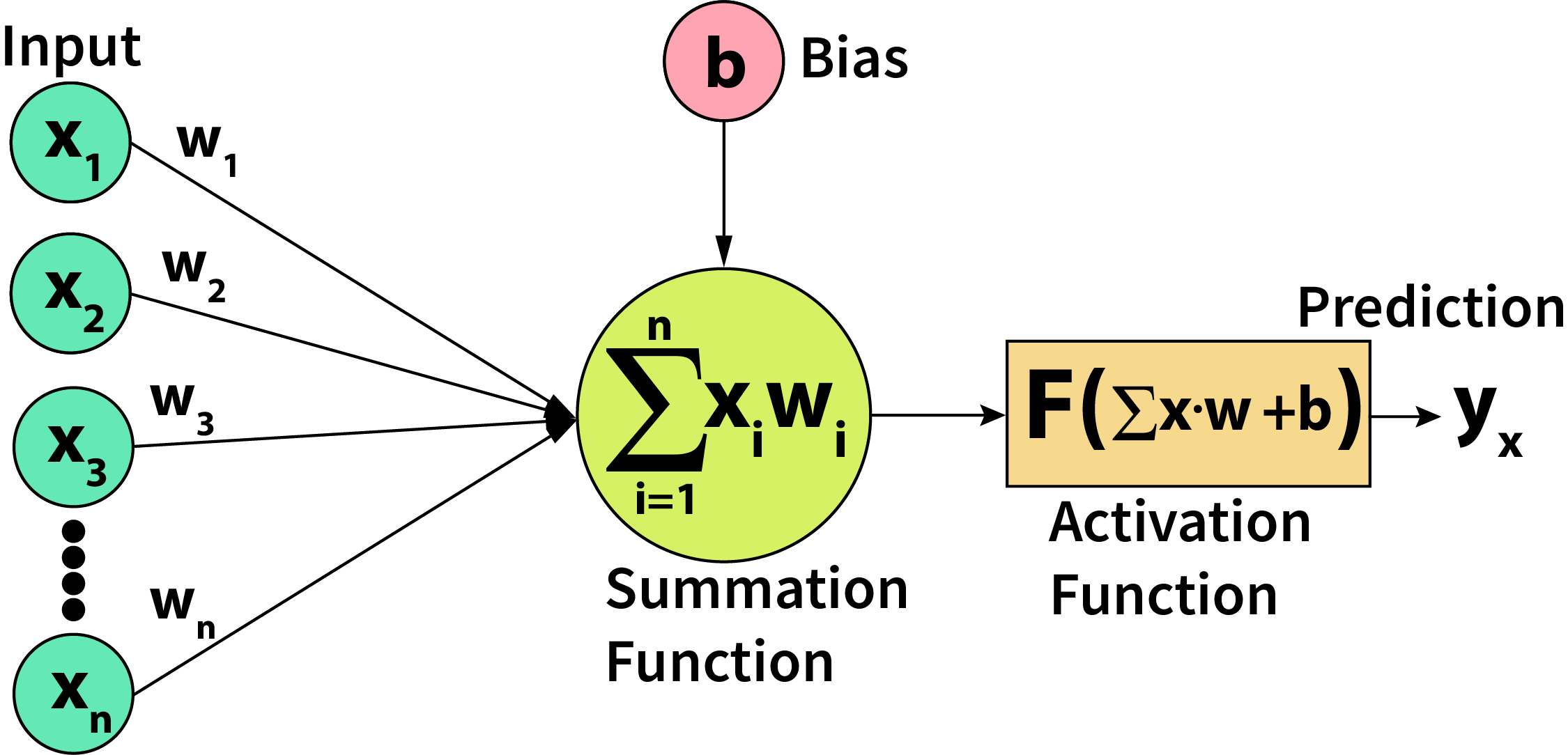}}
\caption{Single Neuron}
\end{figure}

\subsection{Input}

A basic Convolutional Neural Network use image matrices as the input. Images are stored in a system in the form of mathematical matrices. For Example, A colored image of dimensions 1920 $\times$ 1080 is stored in the system as a 3d matrix of size (1920, 1080, 3) where, 1980 is width, 1080 is height and 3 represents the number of color channels i.e. RGB. Each cell in the image matrix contains the RGB value of the corresponding original image as shown in Fig 3. The image matrix is given as an input in the neural network for further processing which involves steps such as feature extraction using convolution and pooling for faster processing and other similar processes.

\begin{figure}[htpb]
\centerline{\hspace{0.3cm}\includegraphics[scale = 1]{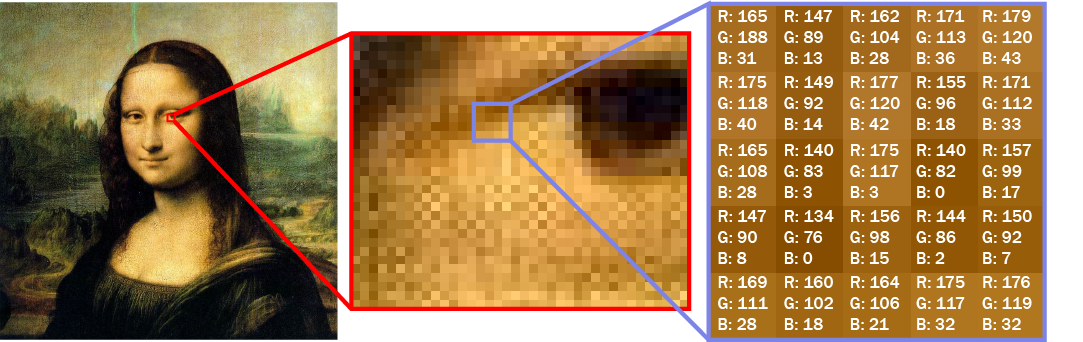}}
\caption{Matrix representation of a color image}
\end{figure}

\subsection{Convolution}

The mathematical process known as convolution is performed on two functions in order to generate a third function that expresses how the shape of one function affects the shape of the second function. In other terms, "convolution" refers to the act of multiplying two functions point-wise to create a third function. In this case, one of the functions is the image pixels matrix, and the other is the filter. After a convolution operation has been performed on an input image matrix, a filter is a tiny matrix that is used to extract features from the input image matrix. Feature Maps is the name given to the matrix that is produced.

In the mathematical form, 

$y\left[i,j\right]=\sum_{m=-\infty}^\infty\sum_{n=-\infty}^\infty h\left[m,n\right] \cdot x\left[i-m,j-n\right]$

where m = No. of rows, n = No. of columns, 
\vspace{0.4cm}

The convolution process for matrices is shown in Fig 4.

\begin{figure}[htpb]
\centerline{\includegraphics[width=7cm, height=5cm]{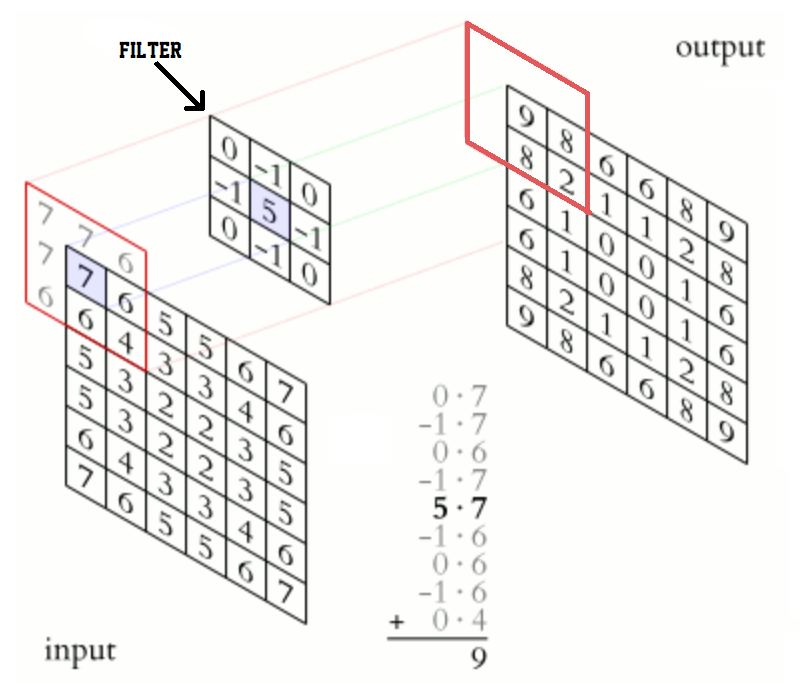}}
\caption{Convolution Operation on Matrices}
\end{figure}

\subsection{Pooling}

The process of pooling entails sliding a small two-dimensional matrix over the feature map in effort to reduce the dimensions of the map without losing the knowledge of the features that are located in that particular location. Pooling reduces the quantity of parameters that must be learned, which in turn makes the calculation more effective. Pooling can be broken down into two distinct categories: maximum pooling and average pooling. The output of the max pooling method is determined by the value that is highest in that region, whereas the output of the average pooling method is determined by the average of all the values in that region, as illustrated in Figure 5.

\begin{figure}[htpb]
\centerline{\includegraphics[scale = .4]{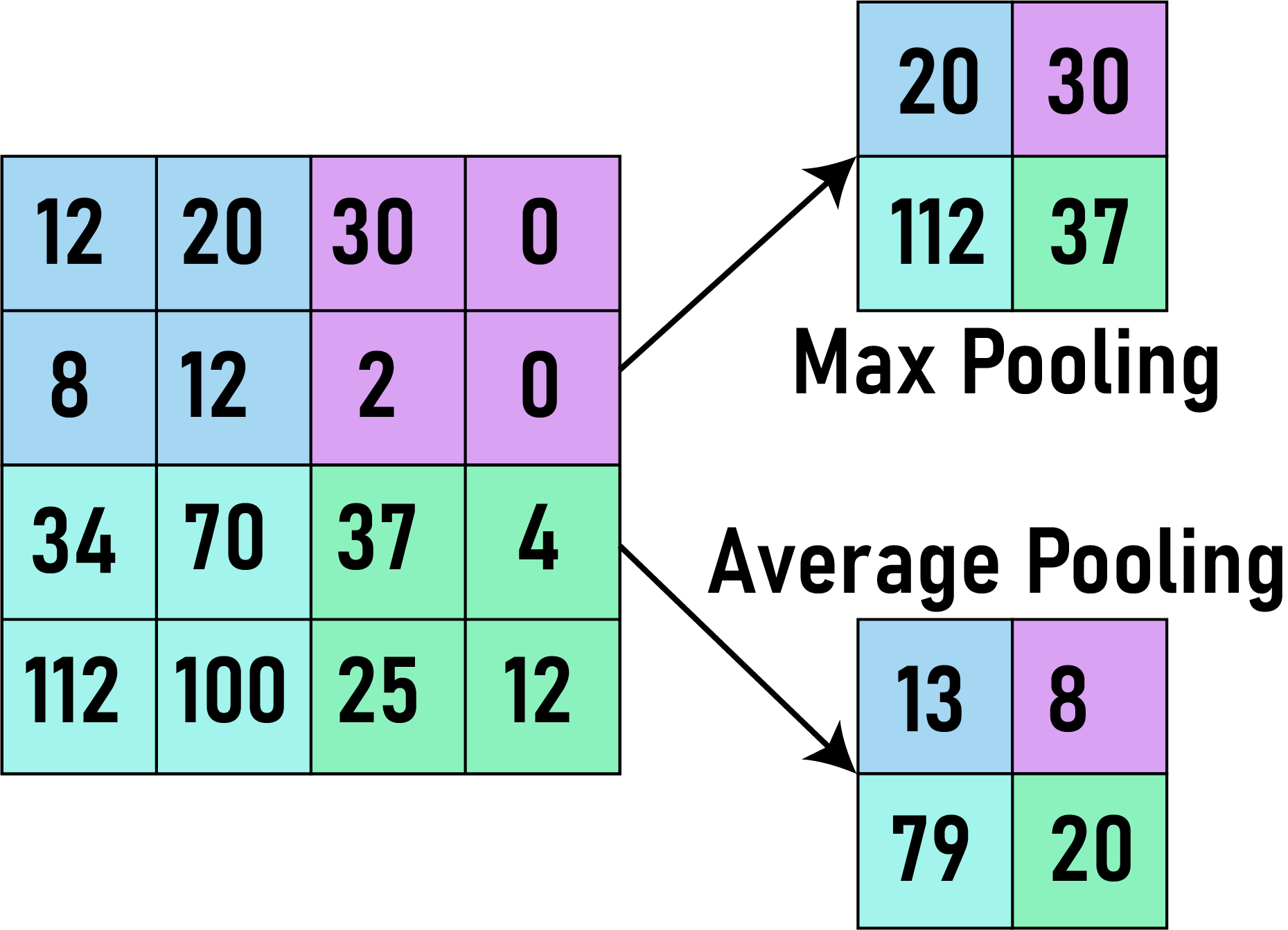}}
\caption{Pooling process}
\end{figure}

\subsection{Regularization}

Regularization is a technique used to counter the problem of decreasing the complexity of the neural network during training to avoid overfitting a model. There are three types of regularization are used L1, L2 and Dropout. We are using the Dropout regularization method which drops or turns off random nodes in the network according to some probability \emph{P}. Thus reducing the complexity of the neural network avoiding overfitting. The Dropout process is given in the Fig 6.

\begin{figure}[htpb]
\centerline{\includegraphics[scale = .4]{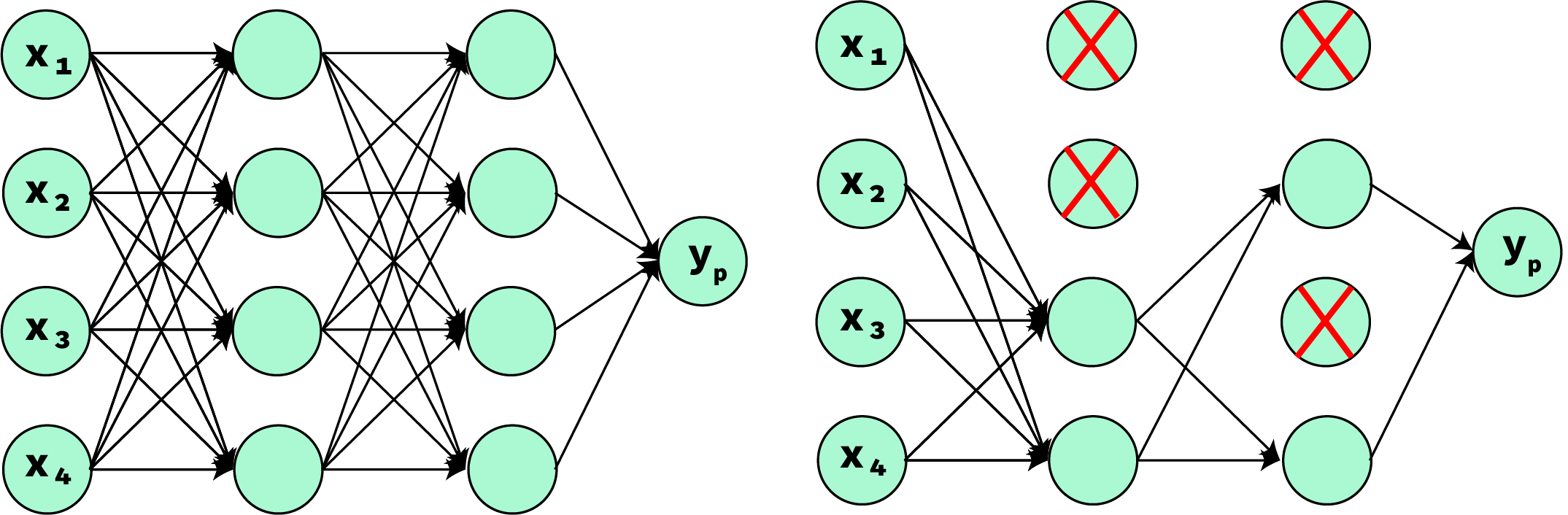}}
\caption{Dropout Regularization with \emph{P}=0.5}
\end{figure}

\subsection{Hidden and Output layers}

Hidden layers also called Fully Connected Layers uses non-linear activation function to apply non-linear transformation on a flattened input i.e. 1-D input. It uses activation functions like Relu (Rectified Linear Unit), tanh, sigmoid, etc. The output layer uses probabilistic functions like Sigmoid or logistic, Softmax, linear to find probability for each class the network is classifying


\section{System Design}

\subsection{Dataset}

The dataset utilized in this project has been contributed by NIST (National Institute of Standards and Technology) which contains a total of 1,01,784 images for a total of 47 classes. The 47 classes include0-9, A-Z, a, b, d, e, f, g, h, n, q, r, t making a total of 47. Few of the small alphabets are considered due to the similarity between small and capital letters which will reduce the complexity for the model training. A sample from the dataset in shown in the below Figure.

\begin{figure}[htpb]
\centerline{\includegraphics[scale=.5]{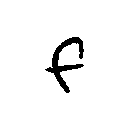}}
\caption{A sample from the character dataset} 
\end{figure}

\subsection{Preprocessing}

The images from the dataset are of dimensions 128 x 128 which is will increase the number of parameters during training exponentially resulting in longer training time. To avoid this issue, the images are rescaled to 32 x 32 and the images along with the labels are converted into two NumPy arrays respectively. The image will use less CPU and GPU resources when resized because there are fewer pixels to process. The images are already in grey-scale so we don't need to add extra filters. The resulting dimensions for a single image is 32 x 32 x 1, where 1 represents the black and white color channel.

\subsection{Splitting into Training and Testing Dataset}

Training and testing datasets are created from the dataset. An optimal ratio of 70:30 is employed for splitting. The training dataset contains 71,249 images and the testing dataset contains 30,535 images. Both the datasets are shuffled using a permutation to reduce variance.

\subsection{CNN Model}

\begin{figure}[htpb]
\centerline{\includegraphics[width=9cm, height=3cm]{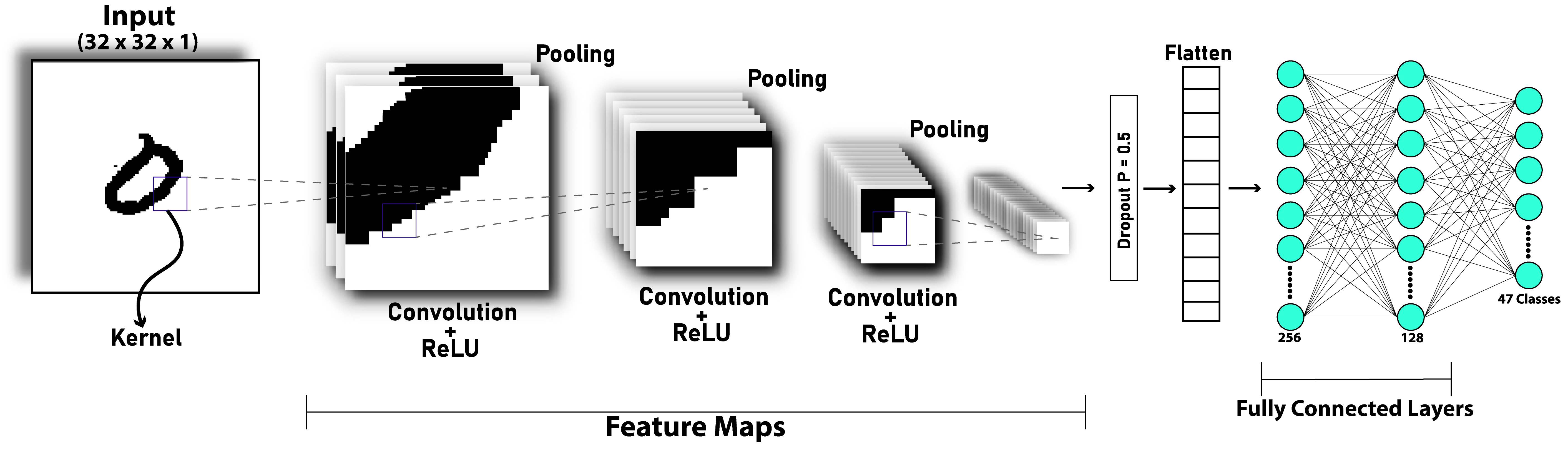}}
\caption{The CNN model architecture}
\label{rem}
\end{figure}

The CNN model used for this project is quite different from the other pre-trained CNN model architecture available like ResNet, GoogleNet, etc. The CNN model used for the experiment is a implemented on a custom built architecture which is displayed in the figure \ref{rem}.

Three Convolution and Pooling layers are used. First conv layer a total of 1024 filters of size 5 x 5. Second conv layer uses 512 filters of size 3 x 3 and Third conv layer uses 256 filters of size 3 x 3. The activation function used for these layers is ReLU(Rectified Linear Unit) which has a range from $ 0 -\infty$. A dropout regularisation layer and a flatten layer are added. In the fully connected layer, two layers containing 256 and 128 neurons respectively connected to the 47 output neurons.

\subsection{Output Layer}

In order to provide an accurate prediction regarding the image's category, the output layer makes use of the softmax probabilistic function. Along with the Adam optimizer, the usage of categorical cross entropy as a loss function is also common. In this case, an active learning algorithm known as the Adam Optimization method is applied. This approach employs individualised learning rates for every one of the parameters.


\subsection{Train the Model}

The CNN model is trained over a period of 20 epochs, with 357 steps each epoch. Figures 9 and 10 shows both the training accuracy against validation accuracy and training loss against validation loss using plots.
\begin{figure}[htbp]
\vspace{0.1cm}
\centerline{\includegraphics[scale=.5]{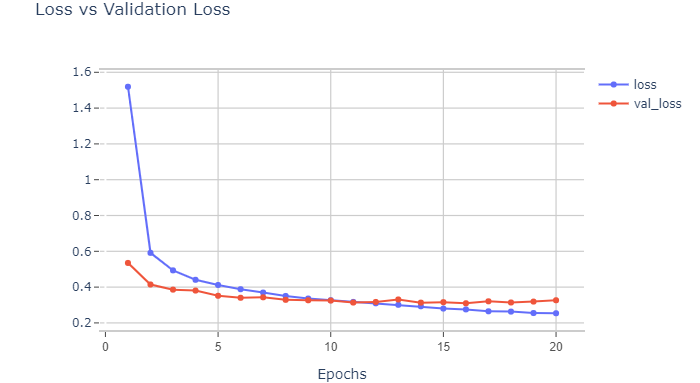}}
\vspace{0.1cm}
\centerline{\includegraphics[scale=.5]{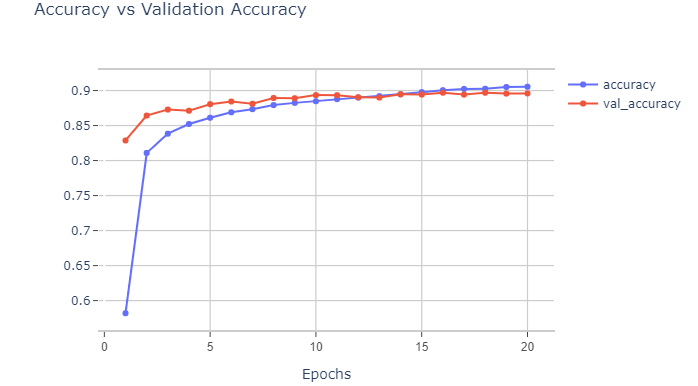}}
\vspace{-0.1cm}
\caption{The line plot about loss and accuracy}
\end{figure}

\section{Results}

The use of a convolutional neural network model to detect handwritten English alphabetic and numeric characters is discussed in this research. Though many advanced techniques has been invented for this problem, CNN being the predecessor of these techniques has given satisfying results. The model which has been developed for this paper has given almost accurate predictions with an accuracy percentage of 90.54\% with a loss of 2.53\%. A learning rate of 0.001 is used as a parameter for a better performance.

\begin{figure}[h]
\centerline{\includegraphics[scale=.2]{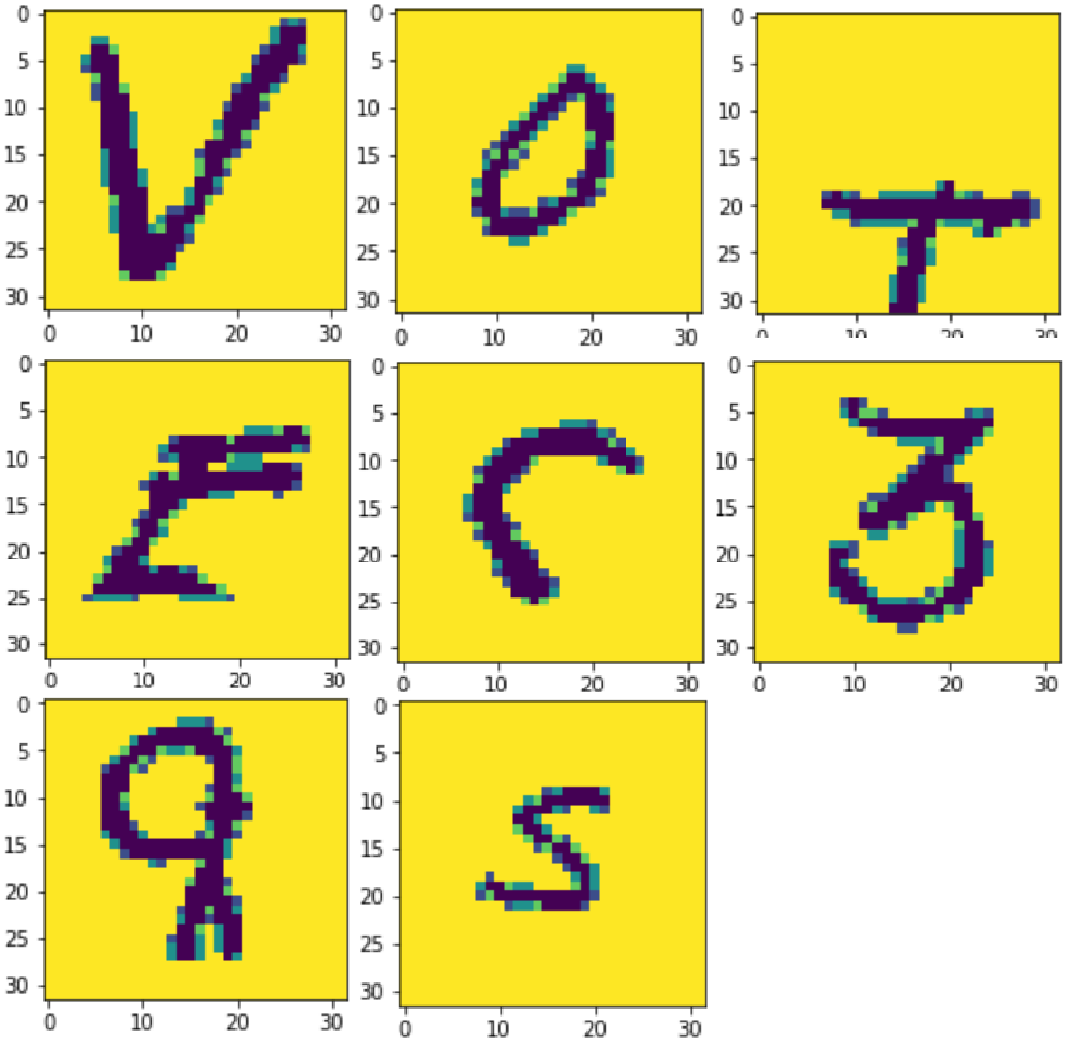}}
\caption{Images for model testing}
\end{figure}

\hspace{2cm}

With a few random photos of the handwritten characters, the model has been evaluated. The sample images used for the testing are shown in the figure 10.The results are shown in the above table. The model has not been perfected yet due to which, a noticeable error can be seen in the table for the class 't' which is  wrongly predicted as 'T'.

\newcolumntype{P}[1]{>{\centering\arraybackslash}p{#1}}
\newcolumntype{M}[1]{>{\centering\arraybackslash}m{#1}}
\begin{table}
  \centering
  \begin{tabular}{|m{3cm}|m{3cm}|}
    \cline{1-2}
    Original Label & Predicted Label \\
    \cline{1-2}
    V & V \\
    \cline{1-2}
    0 & 0 \\
    \cline{1-2}
    t & F \\
    \cline{1-2}
    E & E \\
    \cline{1-2}
    r & r \\
    \cline{1-2}
    S & S \\
    \cline{1-2}
    3 & 3 \\
    \cline{1-2}
    q & q \\
    \cline{1-2}
    S & S \\
    \cline{1-2}
\end{tabular}
  \newline\newline
  \caption{Original Label vs Predicted Label}\label{tab1}
\end{table}
\vspace{0.0cm}


The receiver operating characteristic curve (ROC) curve is shown in Figures 11 through 14 for a variety of classes. The performance of the classification model at different thresholds is depicted on a graph called the ROC curve, or receiver operating characteristic curve. The True Positive Rate, or recall, and False Positive Rate, or precision, are used in this curve.

\hspace{2cm}$TPR = \frac{TP} {TP + FN}$

\vspace{0.3cm}

\hspace{2cm}$FPR = \frac{FP} {FP + TN}$, where TP = True positive

\hspace{5cm}FP = False Positive

\hspace{5cm}FN = False Negative

\vspace{0.4cm}

\begin{figure}
\vspace{-0.2cm}
\centerline{\includegraphics[scale = 0.16]{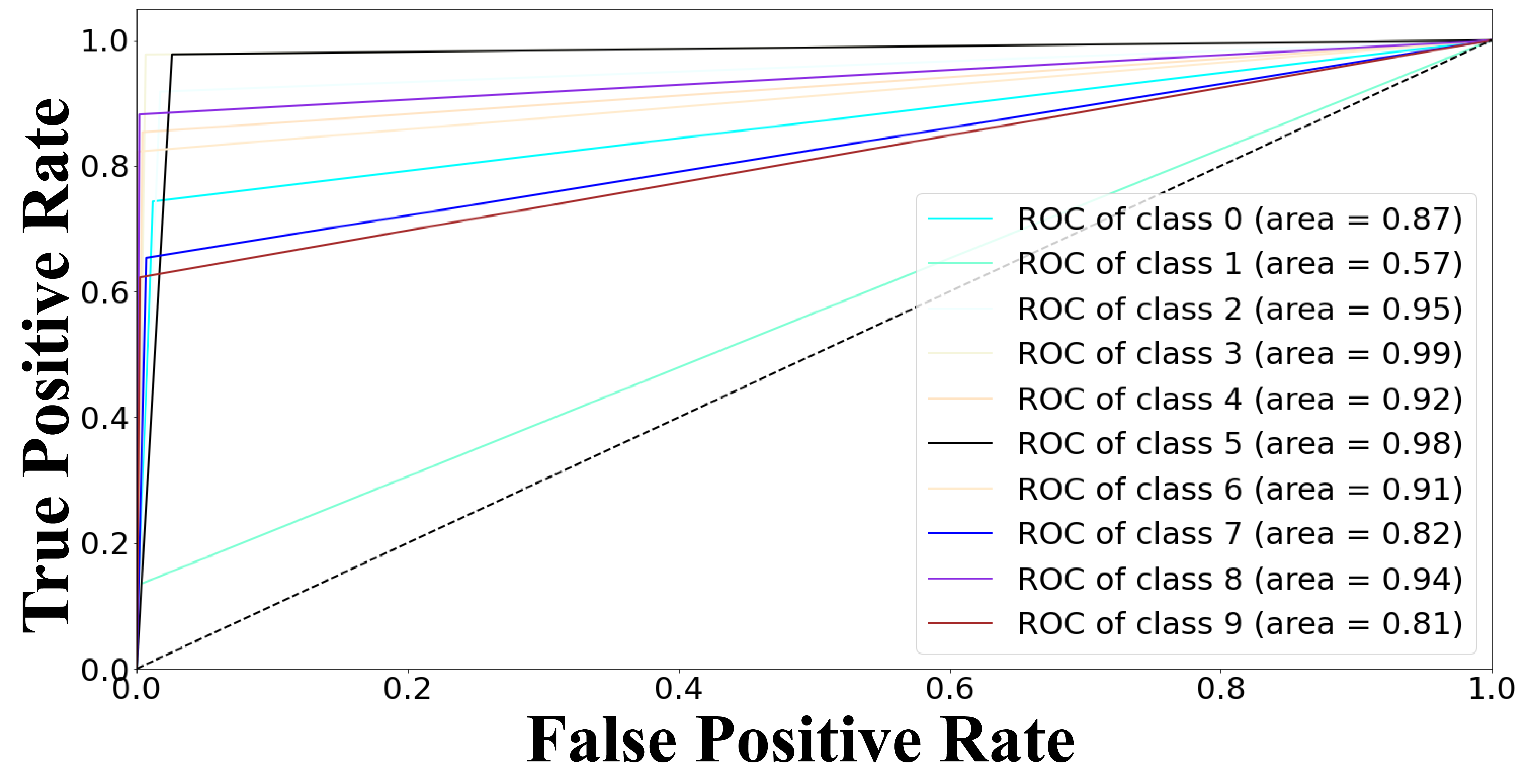}}
\caption{ROC Curve for classes(1-10)}
\vspace{0.3cm}
\centerline{\includegraphics[scale = 0.16]{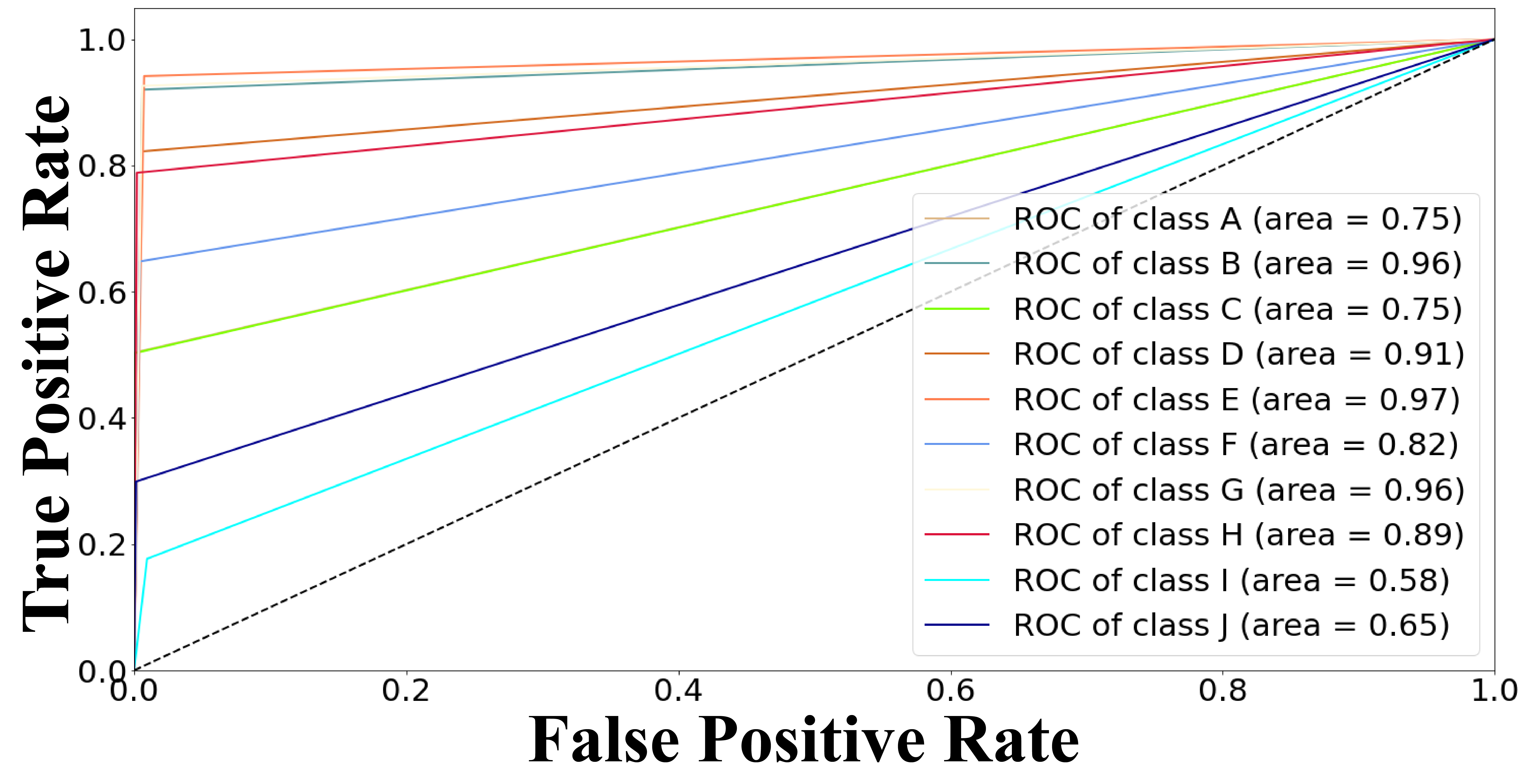}}
\caption{ROC Curve for classes(10-20)}
\vspace{0.3cm}
\centerline{\includegraphics[scale = 0.16]{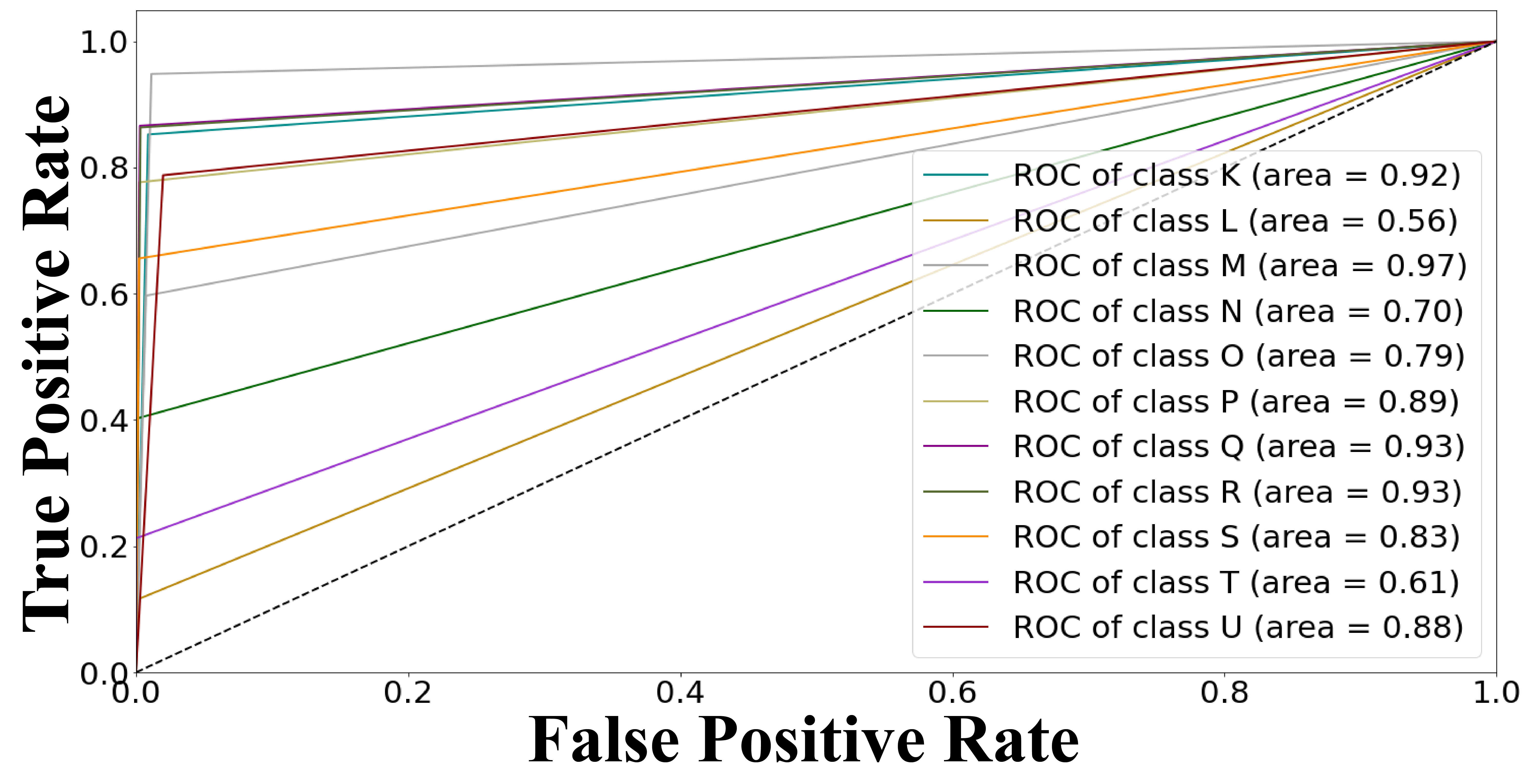}}
\caption{ROC Curve for classes(20-31)}
\vspace{0.3cm}
\centerline{\includegraphics[scale = 0.16]{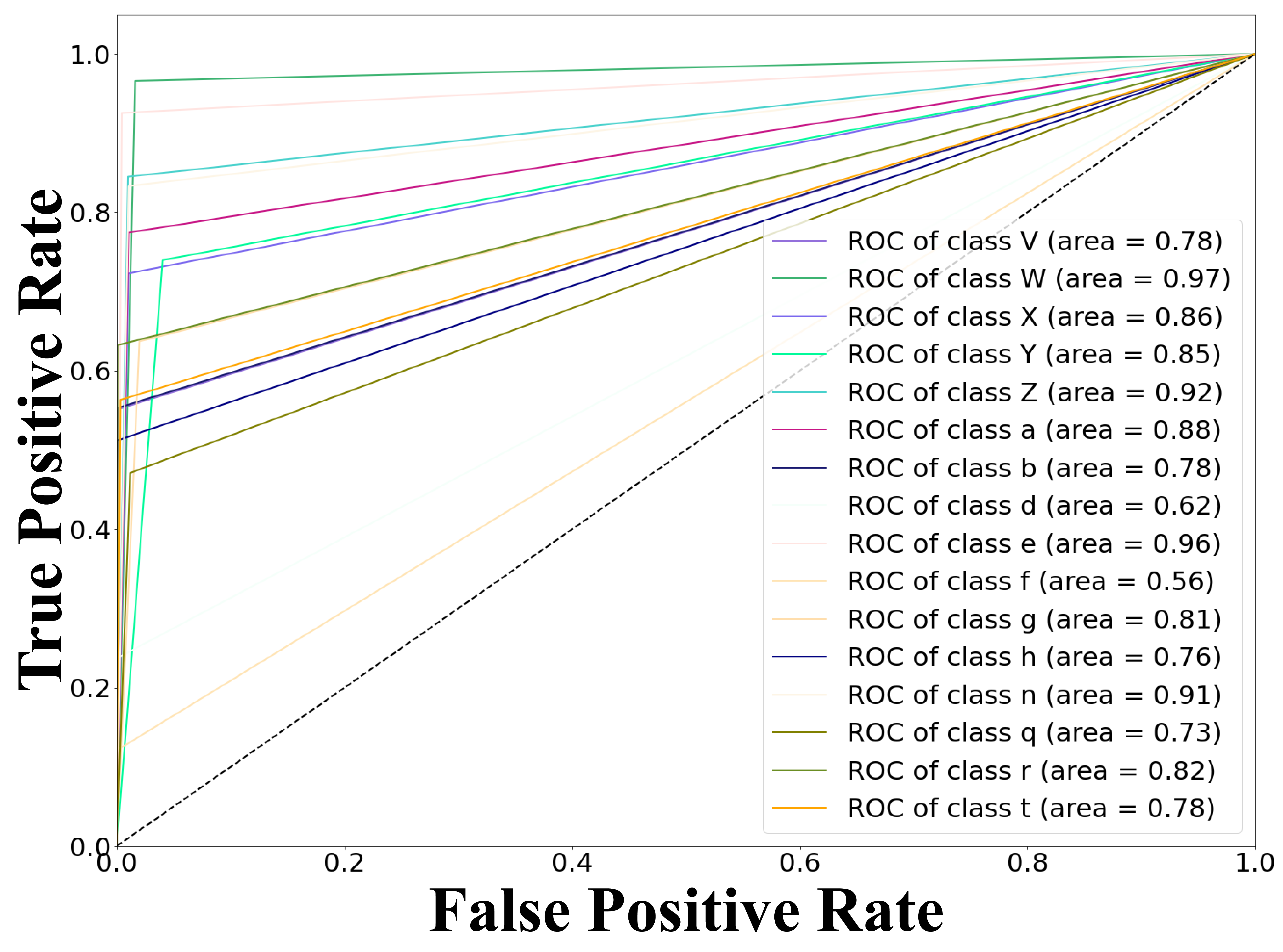}}
\caption{ROC Curve for classes(31-47)}
\end{figure}

The term "precision" refers to the percentage of cases that were correctly predicted but did not turn out to be positive. When the risk of obtaining a false positive is more significant than the risk of obtaining a false negative, precision might be helpful. The term "recall" refers to the percentage of real positive cases that our model properly predicted. This percentage is expressed as a percentage. It is a helpful statistic in situations where the risk of a false negative is more significant than the risk of a false positive.  When both Precision and Recall are equal, it reaches its maximum value.

The two-dimensional area under the ROC curve is measured by the ROC curve's AUC (Area Under the Curve). The AUC offers an overall performance measurement across all categorization criteria. The classification performance for a given class curve is improved by a curve's AUC. The accuracy of a curve's categorization is improved with a lower AUC value. It is clear from the ROC curve figures that the classes "1," "I," "J," "S," "T," and "f" have low AUCs for their curves. The similarity between these characters might be the cause for this error.

\section{Conclusion}

It is possible to draw the conclusion, based on all of the evaluation metrics presented above, that Convolutional Neural Network is an effective method for solving the problem of handwritten character recognition that is also simple to implement and that produces high levels of accuracy and predictions. Although it might not be the most effective recognition algorithm available, it gets the job done nonetheless.

\end{document}